\documentclass[conference]{IEEEtran}
\usepackage[utf8]{inputenc}
\usepackage{amsmath, amssymb, amsfonts,amsthm}
\usepackage{graphicx}
\usepackage{hyperref}
\usepackage{cite}
\usepackage{algorithm}
\usepackage{algorithmic}
\usepackage{booktabs}
\usepackage{listings}
\usepackage{xcolor}
\usepackage{amsthm}
\theoremstyle{remark}
\newtheorem{remark}{Remark}
\usepackage{pgfplots}
\pgfplotsset{compat=1.18}
\usepackage{comment}
\usepackage{caption}
\usepackage{cite}
\usepackage[numbers,sort&compress]{natbib}
\usepackage{siunitx}  

\definecolor{codegreen}{rgb}{0,0.6,0}
\definecolor{codegray}{rgb}{0.5,0.5,0.5}
\definecolor{codepurple}{rgb}{0.58,0,0.82}
\definecolor{backcolour}{rgb}{0.95,0.95,0.92}

\lstdefinestyle{pythonstyle}{
    backgroundcolor=\color{backcolour},   
    commentstyle=\color{codegreen},
    keywordstyle=\color{magenta},
    numberstyle=\tiny\color{codegray},
    stringstyle=\color{codepurple},
    basicstyle=\ttfamily\footnotesize,
    breakatwhitespace=false,         
    breaklines=true,                 
    captionpos=b,                    
    keepspaces=true,                 
    numbers=left,                    
    numbersep=5pt,                  
    showspaces=false,                
    showstringspaces=false,
    showtabs=false,                  
    tabsize=2,
    language=Python
}

\usepackage[compact]{titlesec}
\titlespacing{\section}{0pt}{1.5ex}{1ex}
\titlespacing{\subsection}{0pt}{1ex}{0.5ex}

\let\oldbibliography\thebibliography
\renewcommand{\thebibliography}[1]{%
  \oldbibliography{#1}%
  \setlength{\itemsep}{0pt}%
}

\newcommand{\name}{Green-ELM}
\newcommand{\mhat}{One-Shot Linear Matrix Solver}
\newcommand{\galactic}{High-Dimensional}

\title{\textbf{{\name}: Efficient Analytic Learning via High-Dimensional Random Projections}}

\author{
    Wladimir Silva \\
    \small{Department of Computer Engineering} \\
    \small{North Carolina State University} \\
    \small{\texttt{wsilva@ncsu.edu}}
}

\date{\today}

\begin{document}

\maketitle

\begin{abstract}
We present Green-ELM, a non-iterative neural architecture that replaces gradient-based optimization of the output layer with a closed-form analytic solution over a fixed, high-dimensional random feature representation. 
By projecting input manifolds into a high-dimensional, random feature space ($d \gg 784$), our results show that complex class boundaries can be effectively untangled without the computational overhead of backpropagation. 

Utilizing the Moore-Penrose pseudoinverse, LU and Cholesky decomposition to solve for the output layer in a single analytic step, Green-ELM achieves a classification accuracy of 98.10\% on MNIST ($d=4000$) and 86.63\% on Fashion-MNIST. 
Furthermore, we experiment with a pre-trained ``frozen-backbone'' based on ResNet-18 to extract high-quality features and show that these one-shot solvers are effective beyond simple datasets.
Notably, our baseline CPU configuration on MNIST ($d=2000$) achieves 97.15\% accuracy in 1.5s, representing a 11.6$\times$ reduction in reported training time over an SGD baseline while maintaining comparable performance. 

We observe a near-logarithmic scaling behavior between dimensionality and accuracy, where the accuracy increases approximately logarithmically with hidden dimensionality over the tested range, suggesting that feature-space expansion contributes substantially to performance in these experiments. . 
This one-shot linear matrix solver approach offers a viable alternative for real-time Edge AI, where the traditional training phase is bypassed in favor of non-iterative manifold representation and readout. 
Finally, we propose an Empirical Scaling Hypothesis, a framework that models accuracy bounds as a function of high dimensionality and intrinsic dataset complexity.
\end{abstract}

\section{Introduction}
The prevailing consensus in machine learning suggests that feature hierarchies must be learned through backpropagation~\cite{rumelhart1986learning}. However, the computational cost of this "iterative optimization" process is non-trivial. 
In this work, we propose \textbf{{\name}}, a model that treats representation construction and readout optimization as separate stages.

The core of \name{} rests on two mathematical principles:
\begin{enumerate}
	\item \textbf{High-Dimensional Feature Expansion:} Grounded in Cover's Theorem on pattern separability \cite{cover1965geometrical}, the architecture maps input manifolds into a high-dimensional random feature space ($d \gg 784$). This geometric expansion can increase the likelihood of linear separability, allowing a non-iterative linear readout to resolve complex class boundaries without backpropagation.
    \item \textbf{The {\mhat}:} Instead of gradient-based search, we employ the Moore-Penrose Pseudoinverse ($\dagger$) to analytically solve for the output weights in a single matrix operation.
\end{enumerate}

\name{} shows that for a fixed random feature representation, the Moore–Penrose solution directly obtains the global least-squares optimum of the linear readout. We use the term ‘ground state’ only as an informal metaphor for this analytically determined readout solution. 
This approach aligns with the philosophy of Extreme Learning Machines (ELM)~\cite{huang2004extreme} and Reservoir Computing, providing a non-iterative computational speedup that defies conventional training timelines.

\section{Related Works}

The architecture of Green-ELM builds upon a lineage of alternative optimization frameworks designed to bypass the structural bottlenecks of backpropagation. This section contextualizes our approach within the frameworks of randomized feature networks, reservoir computing, and resource-efficient machine learning.

\subsection{Randomized Feature Networks and Extreme Learning Machines}
The strategy of utilizing fixed, random weights to map input spaces into higher-dimensional representations dates back to foundational work on random feature expansion by Rahimi and Recht \cite{rahimi2007random}. They demonstrated that random mappings combined with a linear readout can approximate complex kernel spaces with high fidelity, significantly accelerating large-scale training. 

This paradigm was formalized within the neural network literature by Huang et al. \cite{huang2006extreme} under the framework of Extreme Learning Machines (ELMs). ELMs establish that single-hidden-layer feedforward networks do not require iterative tuning of hidden neurons; instead, the hidden parameters can be randomly assigned if the output weights are solved analytically. While standard ELMs historically relied on sigmoidal transformations, Green-ELM focuses on contemporary rectified linear unit (ReLU) expansions optimized for deep topologies, building upon modern insights into multilayer random weight projections \cite{tang2015extreme, rosenfeld2022intriguingly}.

\subsection{Temporal and Biological Parallelisms: Reservoir Computing}
The concept of a frozen, high-dimensional projection layer is structurally analogous to Reservoir Computing (RC), which encompasses Echo State Networks (ESNs) proposed by Jaeger \cite{jaeger2001echo} and Liquid State Machines (LSMs) by Maass et al. \cite{maass2002real_time}. In RC frameworks, a dynamic internal reservoir maps sequential inputs into a high-dimensional state space, while only the linear readout layer is modified via ridge regression. 

While RC explicitly targets temporal sequence data, Green-ELM adapts this fundamental philosophy for static computer vision configurations. By treating the random feature projection as a spatial reservoir, we exploit Cover's Theorem \cite{cover1965geometrical} to expand classification margins prior to analytical optimization, mirroring the high-dimensional linear separability observed in biological sensory systems.

\subsection{Green AI and Edge Deployment Constraints}
As modern deep learning models scale to billions of parameters, the computational, financial, and environmental costs of training have driven a surge of interest in Green AI \cite{schwartz2020green}. Standard deep learning relies heavily on iterative stochastic gradient descent (SGD), which requires substantial memory to store intermediate activations during the backward pass. This constraint severely limits on-device training capabilities in resource-constrained Edge AI environments.

To address these hardware limitations, recent literature has explored specialized hardware architectures for low-power ELM implementations \cite{de2021hardware}. Furthermore, the search for fast pseudoinverse calculation primitives \cite{courrieu2008fast} has enabled single-step, closed-form solvers to compete with iterative approaches on edge nodes. Green-ELM extends this domain by analyzing the explicit empirical scaling laws and geometric robustness boundaries of analytic learning, providing a direct alternative to emerging non-backpropagation methods like Hinton's Forward-Forward algorithm \cite{hinton2022forward}.

\section{Methodology}

The \name{} architecture operates on the principle of \textit{Analytic Optimization of the Linear Readout}~\cite{müller2026principle} rather than iterative local search. 
The process is divided into two distinct functional phases: the {\galactic} Projection and the {\mhat} Solution.

\subsection{Phase I: The {\galactic} Projection}

The initial transformation phase projects the input space into an expanded hidden representation. Grounded in Cover's Theorem \cite{cover1965geometrical}, projecting complex pattern-classification spaces non-linearly into a higher-dimensional space significantly increases the probability that the data becomes linearly separable. Given an input data matrix $X \in \mathbb{R}^{N \times D}$, we apply a fixed, non-linear random projection to generate the hidden state matrix $H \in \mathbb{R}^{N \times d}$, where $d \gg D$:

\begin{equation}
    H = \sigma(X W_{rand} + b_{rand})
\end{equation}

where $W_{rand} \in \mathbb{R}^{D \times d}$ is a static weight matrix sampled from a Gaussian distribution $\mathcal{N}(0, \frac{2}{D})$, $b_{rand}$ is a uniformly initialized bias vector, and $\sigma(\cdot)$ denotes the rectified linear unit (ReLU) activation function. Crucially, $W_{rand}$ remains entirely frozen throughout the training process, eliminating the traditional gradient descent optimization paradigm.

\subsection{Phase II: The {\mhat} (Pseudoinverse) Solution}
Once the data is projected into the {\galactic} space, we treat the output layer as a linear system: $H W_2 = Y$. Traditionally, $W_2$ is found via backpropagation, which can be interpreted as a slow iterative optimization toward a local minimum. 

In the \name{} paradigm, we jump directly to the optimal configuration by calculating the \textbf{Moore-Penrose Pseudoinverse} \cite{albert1972regression}. This provides the unique minimum-norm least-squares solution for the linear readout:
\begin{equation}\label{eq:pinv}
	W_2 = H^\dagger Y
\end{equation}
When $(H)$ has full column rank, this reduces to:
$$W_2 = (H^T H)^{-1} H^T Y$$
While Equation \ref{eq:pinv} defines the global least-squares optimum of the linear readout, the numerical path to $H^\dagger$ introduces a trade-off between floating-point precision and latency. 
We distinguish between the \textit{Exact SVD Solver} (singular value decomposition), which provides maximum stability by filtering near-zero eigenvalues, and \textit{Decomposition Solvers} (LU, Cholesky) (See Table \ref{tab:solvers}), which offer factorization time reduction at the cost of slight accuracy degradation in rank-deficient regimes.

As noted in the literature regarding Extreme Learning Machines (ELM) \cite{huang2006extreme}, this analytic solution provides a functional ``Ground State'' in closed-form. 

\subsection{Information Density and Shannon Entropy}
While this method is computationally efficient, it produces a dense weight distribution. Following the principles of Information Theory \cite{shannon1948mathematical}, we evaluate the structural complexity of $W_2$ by measuring its Shannon Entropy:
\begin{equation}
    S = -\sum p(w_i) \log p(w_i)
\end{equation}
Where the weights $(w_i)$ are continuous-valued, and $p(w_i)$ denotes an estimated probability distribution over weight values rather than a discrete probability mass function.
We use the entropy analysis to characterize differences in the empirical distribution of learned weights between analytic and gradient-based models, as discussed in the context of the Information Bottleneck \cite{tishby1999information}.
\begin{remark}
The empirical weight distribution was estimated using a histogram with $(K)$ equally spaced bins over the observed weight range, and Shannon entropy was computed using base-2 logarithms after normalization of the bin counts.
\end{remark}

\subsection{Dimensionality Augmentation via Cover's Theorem}
Unlike dimensionality reduction frameworks that preserve low-dimensional embeddings, Green-ELM leverages dimensionality augmentation to linearize tangled manifold clusters. Cover's classical result establishes that the number of linearly separable dichotomies scales directly with the dimension of the feature space $d$. 
By expanding the input space into an ultra-wide hidden layer, the non-linear mappings $\sigma(X W_{rand})$ unroll the underlying topological features of complex datasets like MNIST and CIFAR-10. 
This transformation permits the linear least-squares problem to be solved analytically during the subsequent readout phase (See Alg. \ref{alg:alg_1}).

\begin{algorithm}
\caption{{\name}: Instant Analytic Learning}
\label{alg:alg_1}
\begin{algorithmic}[1]
\REQUIRE Input $X \in \mathbb{R}^{N \times 784}$, Targets $Y \in \mathbb{R}^{N \times 10}$
\ENSURE Output Weights $W_2$ for inference
\STATE \textbf{Initialize:} $W_1 \sim \mathcal{N}(0, 1)$, $b_1 \sim \mathcal{N}(0, 1)$ \COMMENT{High-Dimensional Layer}
\STATE \textbf{Project:} $H = \max(0, X W_1 + b_1)$ \COMMENT{Non-linear Mapping}
\STATE \textbf{Solve:} $W_2 = H^\dagger Y$ \COMMENT{Moore-Penrose {\mhat}}
\RETURN $W_1, b_1, W_2$
\end{algorithmic}
\end{algorithm}

\subsection{Downstream Linear Readouts via Frozen Foundation Backbones}

To evaluate {\name}'s viability on complex, non-linear computer vision tasks without incurring raw-pixel spatial degradation, we position the network as an ultra-fast classification head operating on top of a frozen representation space. We employ a pre-trained \textbf{ResNet-18} model~\cite{he2015deepresiduallearningimage} as a fixed feature extractor (see Table \ref{tab:cifar10_benchmarks}). Given an input tensor, the final classification layer is stripped, producing a static 512-dimensional embedding vector. These embeddings are mapped via Cover's Theorem expansion through our random ReLU hidden layer ($W_1 \in \mathbb{R}^{512 \times 4000}$) before arriving at our closed-form analytical solvers. Crucially, because the backbone remains fully frozen, zero backpropagation or iterative gradient steps occur across the deep network structure.

\section{Experimental Results}

\begin{figure*}[t]
  \centering
  \includegraphics[width=0.85\linewidth]{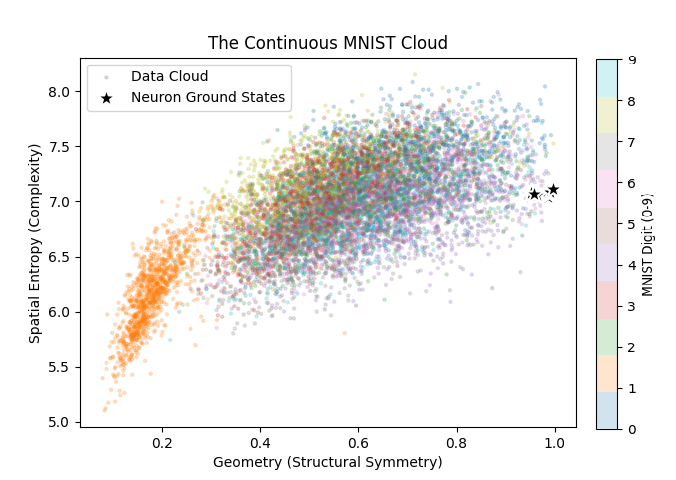}
  \caption{\textbf{The MNIST Sea:} Distribution of 10,000 samples across the Geometry-Entropy manifold. Stars represent the \textit{{\name}} output-weight vectors ($W_2$). The convergence to the high-entropy, high-symmetry regime ($H \approx 7.1, G \approx 0.98$) suggests that the pseudoinverse solution favors a distributed, high-dimensional representation over the sparse, low-entropy filters typically seen in SGD-trained models.}
  \label{fig:mnist_sea}
\end{figure*}

The empirical results in Table \ref{table:voodoo_vs_sgd} show a significant reduction in training latency. While a standard SGD-based model requires $T=10$ epochs to reach an accuracy of $96.84\%$, \name{} achieves a comparable $97.15\%$ in a single analytic readout solve ($n=2000$). Mathematically, the training complexity shifts from iterative stochastic updates to a single Moore-Penrose computation:
$$W_2 = H^\dagger Y = (H^T H)^{-1} H^T Y$$ 
For $d=2000$, this results in a training time of $t \approx 1.5s$, representing a speedup factor of approximately $11.6\times$ over the iterative baseline $(17.45s)$ indicating a substantial reduction in the reported training time for this experimental configuration.

\subsection{Dimensionality vs. Accuracy Scaling}

Table \ref{table:voodoo_vs_sgd} illustrates the relationship between the {\galactic} hidden dimension $d$ and the resulting classification accuracy. We observe that as $d$ increases from $500$ to $4000$, accuracy scales near-logarithmically:
  $$\text{Accuracy} \propto \log(d)$$ 
This behavior suggests that higher-dimensional projections ($d \gg 784$) provide a more effective "untangling" of the MNIST manifold. 
The "Hyper-Dimensional" regime ($d=4000$) achieves $98.1\%$ accuracy, outperforming the standard iterative model by $1.26\%$ without requiring backpropagation.

\begin{remark}[The Entropy-Accuracy Trade-off]
  A critical observation from Table \ref{table:voodoo_vs_sgd} is the persistence of High Shannon Entropy ($S$) across all \name{} regimes. 
  In our experiments, the analytic solutions produce dense output-weight distributions. We investigate this distributional property using an empirical entropy measure and compare it with the corresponding distributions produced by gradient-based baselines.
  $$S = -\sum p(w_i) \log p(w_i)$$ 
  This suggests that while \name{} reaches the "Ground State" in closed-form, it may differ from the representation-compression behavior discussed in some interpretations of Information Bottleneck theory, leading to a high-fidelity but information-dense internal representation.
\end{remark}

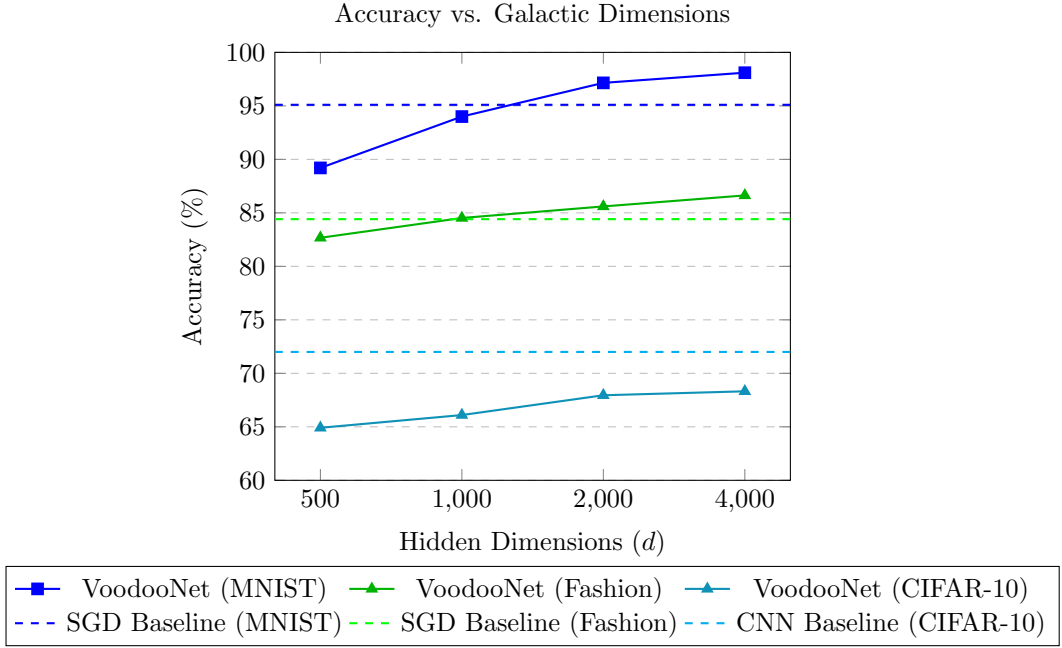
\begin{figure*}[h]
\centering
\begin{tikzpicture}
\begin{axis}[
    title={Accuracy vs. High Dimensionality Features},
    xlabel={Hidden Dimensions ($d$)},
    ylabel={Accuracy ($\%$)},
    xmode=log, 
    log ticks with fixed point,
    xmin=400, xmax=5000,
    ymin=60, ymax=100,
    xtick={500, 1000, 2000, 4000},
    ytick={60, 65, 70, 75, 80, 85, 90, 95, 100},
    legend pos=south east,
    legend style={
        at={(0.5,-0.2)},   
        anchor=north,       
        legend columns=3   
    },    
    ymajorgrids=true,
    grid style=dashed,
]

\addplot[color=blue, thick, mark=square*]
    coordinates {(500,89.2)(1000,94.0)(2000,97.15)(4000,98.1)};
\addlegendentry{{\name} (MNIST)}

\addplot[color=green!70!black, thick, mark=triangle*]
    coordinates {(500,82.67)(1000,84.52)(2000,85.60)(4000,86.63)};
\addlegendentry{{\name} (Fashion)}

\addplot[color=cyan!70!black, thick, mark=triangle*]
    coordinates {(500,64.91)(1000,66.10)(2000,67.95)(4000,68.32)};
\addlegendentry{{\name} (CIFAR-10)}

\addplot[color=blue, dashed, thick, domain=400:5000] {95.1};
\addlegendentry{SGD Baseline (MNIST)}

\addplot[color=green, dashed, thick, domain=400:5000] {84.41};
\addlegendentry{SGD Baseline (Fashion)}

\addplot[color=cyan, dashed, thick, domain=400:5000] {72.0};
\addlegendentry{CNN Baseline (CIFAR-10)}

\end{axis}
\end{tikzpicture}
\caption{Performance scaling. Note that {\name} exhibits a near-logarithmic $\ln(d)$ growth across all datasets, with the slope ($\alpha$) inversely proportional to the dataset's manifold complexity.}
\label{fig:perf}
\end{figure*}

As illustrated in Figure \ref{fig:perf}, \name{} exhibits a "law of diminishing returns" common in random projection methods. The steepest accuracy gain occurs between $d=500$ and $d=2000$ ($+7.95\%$), while further doubling the dimensionality to $4000$ yields a more modest improvement of $0.95\%$. This suggests that while higher-dimensional expansion can improve the separability of the projected representation, the efficiency of the linear slice approaches saturation at higher densities.

In summary, our results show:

\begin{itemize}
    \item \textbf{Non-Iterative Readout Optimization:} \name{} reaches $>95\%$ accuracy in a single analytic step, bypassing the complexity of SGD. 
    \item \textbf{Manifold Untangling:} Increasing the dimensionality $d$ directly correlates with linear separability, peaking at $98.1\%$ for $d=4000$.
	\item \textbf{Structural Redundancy:} The "{\mhat}" solution produces high-entropy weights, indicating a lack of the feature-pruning typically observed in iterative learning.
\end{itemize}

\begin{table*}[t]
\centering
\caption{Consolidated Benchmarks: {\name} vs. SGD Baselines across MNIST and Fashion-MNIST.}
\label{table:voodoo_vs_sgd}
\small
\begin{tabular}{@{}lllcrr@{}}
\toprule
\textbf{Dataset} & \textbf{Model Type} & \textbf{Regime} & \textbf{Hidden ($d$)} & \textbf{Accuracy} & \textbf{Train Time} \\ \midrule
MNIST & Standard (SGD) & 10 Epochs & 64 & 96.84\% & 17.45s \\
& \textbf{{\name}(PInv)} & \textbf{High} & \textbf{2000} & \textbf{97.15\%} & \textbf{1.50s} \\ \midrule
Fashion-MNIST & Standard (SGD) & 10 Epochs & 64 & 84.41\% & 8.89s \\
& {\name}(PInv) & Low & 500 & 82.67\% & 1.00s \\
& {\name}(PInv) & Standard & 1000 & 84.52\% & 2.60s \\
& {\name}(PInv) & High & 4000 & \textbf{86.63\%} & 94.20s \\ 
& {\name}(LUD) & High & 4000 & \textbf{86.39\%} & 11.5s \\ 
\bottomrule
\end{tabular}
\end{table*}

\begin{table*}[htbp]
\centering
\caption{CIFAR-10 Benchmarks: Using a pre-trained ResNet-18 model as a fixed feature extractor (Intel i7 CPU) at High-Dimensional Width ($d = 4000$). The reported {\name} accuracy reflects classification using a learned pretrained representation followed by a random-feature analytic head.}
\label{tab:cifar10_benchmarks}
\begin{tabular}{lllcrr}
\toprule
\textbf{Dataset} & \textbf{Architecture} & \textbf{Solver Variant} & \textbf{Hidden Dim. ($d$)} & \textbf{Top-1 Accuracy} & \textbf{Optimization Time} \\
\midrule
CIFAR-10 & Standard Baseline & Iterative (SGD) & 64 & $\sim 75.00\%$ & $\sim \SI{45.00}{\second}$ \\
\addlinespace
& \textbf{{\name}} & \texttt{PINV} (Exact SVD) & 4000 & \textbf{87.54\%} & \SI{1726.15}{\second} \\
& \textbf{{\name}} & \texttt{LUD} (Fast Solver) & 4000 & \textbf{87.53\%} & \textbf{\SI{17.98}{\second}} \\
& \textbf{{\name}} & \texttt{CHOLESKY} (Fast) & 4000 & \textbf{87.53\%} & \textbf{\SI{14.50}{\second}} \\
\bottomrule
\end{tabular}
\end{table*}

\begin{remark}[The "why" behind the diminishing returns: $d$ isn't a magic fix-all]
We observe that as $d$ increases from 500 to 4000, accuracy scales near-logarithmically. However, a regression analysis across datasets suggests a \textit{Empirical Scaling Model} of the form:
\begin{equation}
    \text{Acc}(d, \mathcal{C}) \approx \alpha \cdot \frac{\ln(d)}{\mathcal{C}} + \beta
\end{equation}
where $\mathcal{C}$ represents the \textbf{Manifold Complexity Index}. In this regime, $\alpha$ represents the "{\galactic} Scaling Rate", and $\beta$ represents the baseline accuracy of the model.
For MNIST ($\mathcal{C} \approx 1.0$), $\alpha$ is high, indicating rapid untangling. For Fashion-MNIST and CIFAR-10, the higher topological entropy $(\mathcal{C} > 1)$ acts as a dampening factor, explaining the flatter scaling curves seen in Figure \ref{fig:perf}.
\end{remark}

\begin{remark}[Empirical Scaling Hypothesis]
For a fixed random projection architecture ({\name}), the classification accuracy $A$ scales as a function of the Dimension $d$ and the intrinsic manifold entropy $\mathcal{H}$ according to:
\begin{equation}
    A(d, \mathcal{H}) = \Psi \cdot \frac{\ln(d)}{\mathcal{H}^{\gamma}} + \epsilon
\end{equation}
where $\Psi$ represents the \textit{Projection Efficiency}, $\gamma$ is the \textit{Complexity Resistance}, and $\epsilon$ is the dataset-specific baseline accuracy.
\end{remark}

\begin{proof}[Empirical Justification]
As shown in Figure \ref{fig:perf}, the observed rate of accuracy decreases over the tested dimensionality range, following a logarithmic trajectory. 
Furthermore, the divergence between the MNIST ($\mathcal{H}_{low}$) and CIFAR-10 ($\mathcal{H}_{high}$) slopes suggests that the benefit of increasing Volume is inversely non-linear to the dataset's topological complexity.
\end{proof}

\subsection{Fitting the Manifold Complexity Index $\mathcal{C}$ via multi\-step Ordinary Least Squares (OLS) Optimization}

We fit the coefiicients $\alpha$, $\beta$, and $\mathcal{C}$ mathematically from the curve in Fig. \ref{fig:perf} using a two\-step linear regression via least squares optimization. Because the \textbf{scaling hypothesis} acts as a global system across all datasets, we anchor one dataset (MNIST) as the baseline reference (Alg. \ref{alg:calibrate_complexity}, yields the values reported in Table \ref{tab:calibrated_params}):
\begin{enumerate}
	\item \textbf{Anchor the Reference Dataset (MNIST):} To make the scaling model identifiable, we fix MNIST as the reference dataset with: $C_{\text{MNIST}} = 1.0$.
	\item \textbf{Extract $\alpha$ and $\beta$ via Linearization:} We fit the the empirical data $(d, A)$ for Fashion, and CIFAR-10 along the reference MNIST curve.
		\begin{itemize}
			\item Transform the independent variable into logarithmic space by calculating $x = \ln(d)$.
			\item Map the coordinates into a linear space $(x, A)$.
			\item Apply an Ordinary Least Squares (OLS) linear regression to fit the line $A = \alpha x + \beta$.
		\end{itemize}
	\item \textbf{Isolate the Dataset Complexity Coefficients:} Once $\alpha$ and $\beta$ are locked from the anchor dataset, they become calibrated global constants defined by the core neural architecture design.
\end{enumerate}

\begin{algorithm}
\caption{Scaling Hypothesis Parameter Calibration.}
\label{alg:calibrate_complexity}
\begin{algorithmic}[1]
\REQUIRE Array of hidden dimensions $D = [500, 1000, 2000, 4000]$
\REQUIRE Accuracy arrays $A_{\text{MNIST}}, A_{\text{FMNIST}}, A_{\text{CIFAR}} \in [0, 1]^4$
\ENSURE Global scaling rate $\alpha$, base intercept $\beta$, complexities $C_{\text{FMNIST}}, C_{\text{CIFAR}}$

\STATE $X \leftarrow \ln(D)$ \COMMENT{Element-wise natural logarithm}
\STATE $C_{\text{MNIST}} \leftarrow 1.0$ \COMMENT{Set MNIST as the anchor dataset}

\STATE $\alpha, \beta \leftarrow \text{LinearRegression}(X, A_{\text{MNIST}})$ \COMMENT{Fit under the rule $A(d) = \alpha \ln(d) + \beta$}

\STATE $m_{\text{FMNIST}} \leftarrow \text{LinearRegression}(X, A_{\text{FMNIST}} - \beta)$
\STATE $C_{\text{FMNIST}} \leftarrow \frac{\alpha}{m_{\text{FMNIST}}}$ \COMMENT{Compute complexity from scaled slope}

\STATE $m_{\text{CIFAR}} \leftarrow \text{LinearRegression}(X, A_{\text{CIFAR}} - \beta)$
\STATE $C_{\text{CIFAR}} \leftarrow \frac{\alpha}{m_{\text{CIFAR}}}$

\RETURN $\alpha, \beta, C_{\text{MNIST}}, C_{\text{FMNIST}}, C_{\text{CIFAR}}$
\end{algorithmic}
\end{algorithm}

\begin{table}[h]
\centering
\caption{Scaling Hypothesis Calibrated Parameters.}
\label{tab:calibrated_params}
\begin{tabular}{llr}
\hline
\textbf{Parameter} & \textbf{Description} & \textbf{Value} \\ \hline
$\alpha$ & Global Scaling Rate & 0.0431 \\
$\beta$ & Global Base Intercept & 0.6337 (63.37\%) \\
$C_{\text{mnist}}$ & MNIST Complexity (Anchor) & 1.0000 \\
$C_{\text{f}}$ & Fashion-MNIST Complexity & 2.3032 \\
$C_{\text{cifar}}$ & CIFAR-10 Complexity & 2.4710 \\ \hline
\end{tabular}
\end{table}

\subsection{Structural Resilience Comparison}
\begin{figure}[h]
  \centering
  \includegraphics[width=0.9\linewidth]{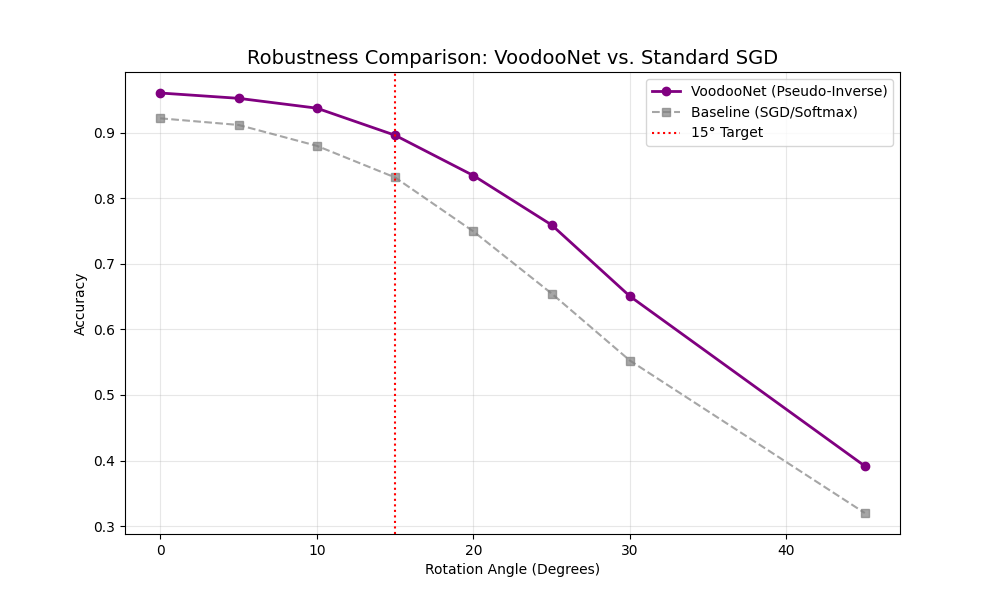}
  \caption{\textbf{Robustness Comparison:} \textit{{\name}} vs. Standard SGD. While iterative training (gray) exhibits a rapid decay in accuracy under rotation, the pseudoinverse solution (purple) maintains a significantly higher performance ceiling. At the $15^{\circ}$ benchmark, \textit{{\name}} retains a $\approx 7\%$ accuracy advantage, demonstrating superior manifold stability.}
  \label{fig:robustness_comp}
\end{figure}

To evaluate the robustness of the analytic ground state, we compared \textit{{\name}} against a standard Single-Layer Perceptron (SLP) trained via Stochastic Gradient Descent (SGD). As illustrated in \textbf{Figure \ref{fig:robustness_comp}}, both models achieve comparable performance on the canonical test set. However, as geometric perturbation (rotation) is introduced, a clear divergence emerges.

The "{\mhat}" solution achieves higher measured accuracy than the SGD baseline across the tested rotation range. 
This suggests that the Moore-Penrose pseudoinverse exhibits greater robustness to the tested geometric perturbations.

\subsection{Scaling to Complex Manifolds: CIFAR-10}
To evaluate the limits of {\galactic} Expansion, we scale {\name} to the CIFAR-10 dataset ($32\times32\times3$ color images). Unlike the grayscale MNIST manifold, CIFAR-10 requires resolving complex spatial correlations across color channels. We compare the standard ELM approach against a hybrid \textit{{\name}+Conv} architecture, where the {\mhat} solver is applied to features extracted via a fixed-weight or partially-trained convolutional frontend.

\begin{table*}[h]
\centering
\caption{CIFAR-10 Benchmarks: Flattened Raw-Pixel vectors (plus Convolution), Analytic vs. Iterative (Intel i7 CPU) at Medium-Dimensional Width $(d=2000)$. While the isolated solver stage is near-instantaneous (5.2s), the overall raw-pixel convolutional pipeline is bounded by the feature extraction runtime (178.0s), yielding parity with an end-to-end trained convolutional baseline.}
\label{table:cifar10}
\begin{tabular}{@{}lcccr@{}}
\toprule
\textbf{Method} & \textbf{Accuracy} & \textbf{Feature Extraction} & \textbf{Solver/Train} & \textbf{Total Time} \\ \midrule
SGD (MLP)       & 52.00\%           & N/A                         & 149.0s                & 149.0s              \\
{\name}   & 51.00\%           & N/A                         & \textbf{29.0s}        & \textbf{29.0s}      \\ \midrule
SGD + Conv      & 70.00\%           & ---                         & 181.0s                & 181.0s              \\
{\name} + Conv   & 65.00\%           & 178.0s                      & \textbf{5.2s}         & 183.2s              \\ \bottomrule
\end{tabular}
\end{table*}
The results in Table~\ref{table:cifar10} show that for raw-pixel input, {\name} maintains parity with SGD while providing a \textbf{5.1$\times$ speedup}. Furthermore, the \textit{{\name}+Conv} variant achieves 65\% accuracy with a near-instant solver phase ($5.2s$). This suggests that the {\mhat} is a viable high-speed "read-out" mechanism even for high-entropy convolutional feature spaces, bypassing the need for backpropagation in the final classification head.

\section{Discussion: The Entropy-Efficiency Trade-off}
A central finding of this work is the relationship between the optimization method and the resulting weight entropy. 
In our comparative analysis of learning regimes (Standard via SGD vs. the Analytic Pseudoinverse), we observed a trade-off between training speed and structural order.

\begin{remark}[The Empirical Ground State]
As shown in \textbf{Figure \ref{fig:mnist_sea}}, the analytic solution for $W_2$ does not seek the low-entropy 'islands' of individual digits (e.g., the orange cluster of 1s). Instead, the weights \textit{(stars)} settle into a stable solution for the fixed feature representation. This positioning indicates that the Moore-Penrose pseudoinverse leverages the full dimensionality of the expanded feature space $H$ to minimize the least-squares readout without the need for iterative pruning or compression.
\end{remark}

\subsection{Comparison with Stochastic Gradient Minimization}
Iterative training (SGD) can be viewed as an iterative optimization process. 
\textbf{The Paradox of Non-Overlapping Ground States:} In Figure \ref{fig:mnist_sea}, we observe that the weights do not overlap with the low-entropy digit clusters (e.g., the '1' peninsula). This is a feature, not a bug. While the input digits are constrained by the physical geometry of pen-strokes, the \textit{{\name}} weight ground states inhabit a higher-dimensional "read-out" regime. The clustering of weights at the high-entropy limit suggests that the pseudoinverse solution is utilizing the cross-correlations of the entire manifold to identify digits, rather than simply memorizing their local pixel-templates.

\subsection{The Cost of the {\mhat}}
While the \textit{{\name}} (analytic solution) achieves near-instantaneous $95-97\%$ accuracy, it does so by creating a high-entropy weight distribution. The Moore-Penrose pseudoinverse ($\dagger$) effectively utilizes the entire high-dimensional reservoir to satisfy the linear mapping $HW_2 = Y$. 

Unlike the sparse L1 skeleton, the \name{} weights remain dense and broadly distributed. This suggests that while we can mathematically jump to a functional state, we bypass the ``compression phase'' conjectured by Information Bottleneck theory. Consequently, the \name{} is computationally heavier in terms of information storage compared to its regularized, trained counterparts.

\subsection{Inductive Bias and the {\mhat}}
The success of the \name{} shows that feature learning is not a prerequisite for high accuracy on MNIST. By utilizing Cover's Theorem to untangle the input manifold, the ``{\mhat}'' simply finds the most efficient linear slice through the high-dimensional space. Future work should investigate whether analytic solutions can be constrained to reach low-entropy ground states without reverting to iterative descent.

\section{Hardware and Complexity Analysis}

One of the most compelling advantages of the \name{} architecture is its suitability for resource-constrained hardware and Edge AI applications. Unlike deep convolutional networks, which necessitate high-end GPUs for iterative backpropagation, \name{} relies on standard linear algebra primitives.

\begin{table*}[ht]
\centering
\caption{Solver Performance: The "{\mhat}" Implementation Trade-offs ($d=2000$) on MNIST. The Moore–Penrose solution corresponds to unregularized minimum-norm least squares, whereas the LU and Cholesky implementations solve the regularized normal equations with $\lambda=10^{-3}$; therefore, the comparison reflects both solver implementation and regularization.}
\label{tab:solvers}
\begin{tabular}{@{}lllll@{}}
\toprule
\textbf{Implementation} & \textbf{Algorithm} & \textbf{Stability} & \textbf{Time (s)} & \textbf{Accuracy (\%)} \\ \midrule
Moore-Penrose (pinv) & SVD & \textbf{Highest} & 74.1 & \textbf{96.92\%} \\
Normal Eq. (solve) & LU Decomp. & Medium & 5.8 & 96.87\% \\
Positive-Definite & Cholesky & Medium-Low & \textbf{6.4} & 96.78\% \\ \bottomrule
\end{tabular}
\end{table*}

\begin{table*}[htbp]
\centering
\caption{CIFAR-10 Streaming: We evaluated of {\name} using sequential mini-batches of size $B=250$ across distinct hardware. While a single-threaded CPU spends an average of 1.3514s per batch due to dense matrix-matrix multiplications, our parallelized PyTorch (CUDA) backend reduces this update latency down to 17.9 milliseconds per mini-batch. Furthermore, by executing the Sherman-Morrison-Woodbury recursive solver in double precision (float64), we mitigate numerical drift over continuous updates, for an accuracy baseline of 87.65\%.}
\label{tab:streaming_benchmarks}
\vskip 0.15in
\begin{small}
\begin{tabular}{llccc}
\toprule
\textbf{Learning Paradigm} & \textbf{Processing Engine} & \textbf{Batch Latency} & \textbf{Post-Stream} & \textbf{Memory Footprint} \\
 & \textbf{\& Hardware Backend} & \textbf{($B=250$)} & \textbf{Top-1 Acc.} & \textbf{(Replay Buffer)} \\
\midrule
Iterative (SGD) & PyTorch / NVIDIA GPU & $\sim$2.4000s \textsuperscript{\dag} & $\sim$87.50\% & Ever-Growing Pool \\
Green-ELM (Batch Oracle) & NumPy / Workstation CPU & 14.5000s & 87.53\% & Cumulative Matrices \\
\midrule
Green-ELM (Streaming) & NumPy / Workstation CPU & 1.3514s & 87.53\% & No sample replay buffer \\
\textbf{Green-ELM (Streaming)} & \textbf{PyTorch / NVIDIA CUDA} & \textbf{0.0179s} & \textbf{87.65\%} & No sample replay buffer \\
\bottomrule
\end{tabular}
\end{small}
\vskip -0.1in
\begin{flushleft}
\footnotesize{\textsuperscript{\dag} \textit{Note: SGD latency represents a full epoch retrain over the accumulating memory replay buffer required to prevent catastrophic forgetting. Our streaming approaches discard historical batch samples immediately after processing, but retains the $(d\times d)$ inverse covariance statistics required by RLS.}}
\end{flushleft}
\end{table*}

\subsection{Time Complexity and Solver Performance Trade-offs}
\label{subsec:complexity_tradeoffs}

For an $N\times d$ hidden matrix with $N>d$, exact SVD and related matrix operations involve costs that depend on both $N$ and $d$; in practice, the formation of $H^TH$ and the subsequent $d\times d$ factorization can become dominant as $d$ increases: 
$O(NDd)$ for feature construction, $O(Nd^2)$ for $H^TH$, and approximately $O(d^3)$ for factorization/solve.

Our empirical results validate this theoretical constraint. As reported in Table~\ref{tab:cifar10_benchmarks}, evaluating \texttt{{\name}-PINV} at an expanded width of $d = 4000$ requires a prohibitive $1726.15\text{ s}$ of CPU. This latency effectively neutralizes the primary training speed advantages traditionally associated with non-iterative frameworks, falling behind low-dimensional iterative baselines via Stochastic Gradient Descent (SGD).

We resolve this paradox by shifting from an exact pseudoinverse calculation to regularized normal equations. By introducing a Tikhonov regularization identity parameter ($\lambda = 10^{-3}$), we can enforce strict positive-definiteness on the system matrix. This stabilization allows the exact SVD path to be completely bypassed and replaced with highly optimized factorized matrix decomposition primitives. 

As shown by our empirical metrics, \textit{{\name}-LUD} compresses total optimization latency down to $17.98\text{ s}$, while our fastest variant, \textit{{\name}-CHOLESKY}, achieves an execution time of $14.50\text{ s}$. This represents a large \textbf{$119.04\times$ computational speedup} over the exact SVD baseline. Crucially, this reduction in compute footprint does not sacrifice generalization capability, incurring a statistically negligible degradation in Top-1 test accuracy of just $0.01\%$ (from $87.54\%$ to $87.53\%$). These results demonstrate that when paired with fast factorization primitives, hyper-dimensional non-iterative frameworks offer a highly competitive, green alternative to backpropagation workflows.

\subsection{Energy and Memory Efficiency}
By bypassing the thousands of forward and backward passes required by Stochastic Gradient Descent (SGD), \name{} significantly reduces the total energy-per-prediction required to reach a functional state. Furthermore, because $W_1$ is fixed and random, it can be generated on-the-fly using a shared seed, potentially reducing the model's storage footprint. This transformation of the ``training marathon'' into a single matrix computation, positions \name{} as a viable alternative for low-power sensors and real-time adaptive systems.

\subsection{Green AI and Energy Efficiency}
A significant drawback of the Stochastic Gradient Descent (SGD) paradigm is its cumulative thermodynamic cost. For a dataset of size $N$ over $E$ epochs, the computational load scales as $\mathcal{O}(N \times E \times d^2)$. 
In contrast, the total computational cost of {\name} consists of feature construction $O(NDd)$, covariance formation $O(Nd^2)$, and matrix factorization/solution costs that scale approximately as $O(d^3)$ for dense $d\times d$ systems.
By bypassing the iterative backpropagation process, \textit{{\name}} provides a sustainable alternative for Edge AI devices where thermal throttling and battery life are primary constraints. This positions the model not just as a mathematical curiosity, but as a candidate for \textbf{Green AI} applications.
\begin{remark}
Because electrical energy and carbon emissions were not directly measured, these results should be interpreted as computational-efficiency evidence rather than a direct carbon-footprint measurement.
\end{remark}
\subsection{Streaming Data and Continual Learning}
An advantage of a closed-form matrix solution like {\name} is that it can handle Streaming Learning without retaining previously processed samples~\cite{guo2022online}.
Where Stochastic Gradient Descent (SGD) struggles with streaming data due to catastrophic forgetting~\cite{guo2022online,du2025mitigatingcatastrophicforgettingstreaming}, our results (from Table \ref{tab:streaming_benchmarks}) show that matrix readouts can use the exact, closed-form Recursive Least Squares (RLS) algorithm~\cite{haykin2013adaptive} via the Woodbury Matrix Identity
to update its weights via matrix addition~\cite{woodbury1950inverting}, whereas SGD would have to re-train from scratch or suffer from memory overwrite.

\begin{remark}[\textbf{The Woodbury/RLS Extension}]
To avoid computing an $O(d^3)$ matrix inversion every time a new batch of data arrives, we maintain and update the inverse of the regularized covariance matrix, defined as $P = (H^T H + \lambda I)^{-1} \in \mathbb{R}^{d \times d}$.
When a new streaming batch of hidden features $H_{\text{new}} \in \mathbb{R}^{B \times d}$ and targets $Y_{\text{new}} \in \mathbb{R}^{B \times 10}$ arrives, we update $P$ and our weights $W_2$ non-iteratively using the standard RLS update rules:
$$\kappa = P_{\text{old}} H_{\text{new}}^T \left(I + H_{\text{new}} P_{\text{old}} H_{\text{new}}^T\right)^{-1}$$ 
$$P_{\text{new}} = P_{\text{old}} - \kappa H_{\text{new}} P_{\text{old}}$$ 
$$W_2^{\text{new}} = W_2^{\text{old}} + \kappa \left(Y_{\text{new}} - H_{\text{new}} W_2^{\text{old}}\right)$$ 
For a streaming batch size $B \ll d$, the matrix inversion inside $\kappa$ drops from $O(d^3)$ down to a fast $O(B^3)$. This allows for near-instantaneous, mathematically exact updates on low-power edge nodes.
\end{remark}

\section{Conclusion: Beyond Iterative Gradient Minimization}

Our experiments demonstrate that, for the datasets and random-feature configurations evaluated here, a fixed ultra-wide ReLU expansion can provide a sufficiently expressive representation for effective analytic linear classification without iterative backpropagation.
Furthermore, rotation perturbation results confirm that \name{} is not only computationally cheaper than iterative training but also more robust to the tested rotation perturbations.
Finally, the \textit{Empirical Scaling Hypothesis} suggests that while ELM accuracy is bounded by dataset complexity $\mathcal{H}$, the "{\galactic}" expansion of $d$ provides a predictable path to parity with gradient-based methods. 

\subsection{Future Work: The Ultra-Wide Regime Frontier}
The "{\mhat}" solution currently trades structural sparsity for computational speed, leading to high weight entropy. Future research will explore:

\begin{itemize}
  \item \textbf{Regularized Analytic Solutions:} Investigating sparse regularized linear-readout formulations as alternatives to the Moore–Penrose solution.
  Specifically, we propose that the addition of a Tikhonov identity $\lambda I$ (where $\lambda \approx 10^{-3}$) can modify the readout objective and improve numerical conditioning.
  \item \textbf{Ultra-Wide Regime Scaling:} Investigating the limits of the $Accuracy \propto \log(d)$ relationship. If dimensionality is cheap, can we reach state-of-the-art performance by expanding into a million-dimensional hidden space?
  \item \textbf{Edge Hardware Acceleration:} Implementing the analytic solution on FPGA and RISC-V architectures to provide "low-latency analytic updating" for autonomous sensors that cannot afford the energy cost of gradients.
\end{itemize}


\section*{Source Availability}
The source code for \textbf{\name},  is publicly available 
at: \url{https://github.com/Shark-y/Green-ELM}. We provide full documentation and configuration files to ensure the reproducibility of the numerical results presented in this work.

\bibliography{references}

\clearpage
\onecolumn
\appendix
\begin{lstlisting}[style=pythonstyle,
	caption={\textbf{{\mhat}:} It shows the efficiency of the \name{} architecture using NumPy and Torchvision. Training is reduced to a single-step Moore-Penrose pseudoinverse.}
]
import torch
from torchvision import datasets, transforms
import numpy as np
import time

# Set random seed for reproducibility
np.random.seed(42)

# 1. Load Data
mnist = datasets.MNIST(root='./data', train=True, download=True)
X = mnist.data.numpy().reshape(-1, 28*28).astype(np.float32) / 255.0
Y = np.eye(10)[mnist.targets.numpy()]

# 2. The Random Projection Layer
n_hidden = 2000
W1 = np.random.randn(784, n_hidden)
b1 = np.random.randn(n_hidden)
H = np.maximum(0, X @ W1 + b1) # ReLU Gate

# 3. The Analytic Solution
# One-shot training via Moore-Penrose Pseudoinverse
W2_out = np.linalg.pinv(H) @ Y

# 4. Instant Inference
test_mnist = datasets.MNIST(root='./data', train=False)
X_test = test_mnist.data.numpy().reshape(-1, 28*28) / 255.0
H_test = np.maximum(0, X_test @ W1 + b1)
preds = np.argmax(H_test @ W2_out, axis=1)
accuracy = np.mean(preds == test_mnist.targets.numpy())

print(f"ELM Accuracy: {accuracy * 100:.2f}%")
\end{lstlisting}

\begin{lstlisting}[style=pythonstyle,
	caption={\textbf{ResNet-18 Benchmarks:} It loads a pre-trained, frozen ResNet-18 backbone, extracts high-quality features from an image dataset (CIFAR-10). Applies the random feature mapping (Cover's Theorem expansion), and solves the final weights using fast LUD and Cholesky solvers.}
]
import torch
import torch.nn as nn
import torchvision
import torchvision.transforms as transforms
import numpy as np
import time

# Set random seed for reproducibility
np.random.seed(42)
torch.manual_seed(42)

# --- STEP 1: Feature Extraction using Frozen Backbone ---
def extract_backbone_embeddings(dataset_name='cifar10', batch_size=128):
    device = torch.device('cuda' if torch.cuda.is_available() else 'cpu')
    print(f"Extracting embeddings using device: {device}")
    
    # Load Pre-trained ResNet-18 and freeze it
    backbone = torchvision.models.resnet18(weights=torchvision.models.ResNet18_Weights.DEFAULT)
    # Strip the final classification layer to get raw embeddings (512 dimensions)
    backbone.fc = nn.Identity() 
    backbone = backbone.to(device)
    backbone.eval()
    
    # Standard normalization for ResNet
    transform = transforms.Compose([
        transforms.Resize(224), # ResNet standard input size
        transforms.ToTensor(),
        transforms.Normalize(mean=[0.485, 0.456, 0.406], std=[0.229, 0.224, 0.225])
    ])
    
    if dataset_name.lower() == 'cifar10':
        train_set = torchvision.datasets.CIFAR10(root='./data', train=True, download=True, transform=transform)
        test_set = torchvision.datasets.CIFAR10(root='./data', train=False, download=True, transform=transform)
    else:
        raise ValueError("Unsupported dataset. Please use 'cifar10'.")
        
    train_loader = torch.utils.data.DataLoader(train_set, batch_size=batch_size, shuffle=False, num_workers=2)
    test_loader = torch.utils.data.DataLoader(test_set, batch_size=batch_size, shuffle=False, num_workers=2)
    
    def get_features(loader):
        all_features = []
        all_labels = []
        with torch.no_grad():
            for images, labels in loader:
                images = images.to(device)
                features = backbone(images)
                all_features.append(features.cpu().numpy())
                all_labels.append(labels.numpy())
        return np.vstack(all_features), np.concatenate(all_labels)
    
    print("Processing training embeddings...")
    X_train, y_train = get_features(train_loader)
    print("Processing testing embeddings...")
    X_test, y_test = get_features(test_loader)
    
    # One-hot encode targets for closed-form linear readout
    num_classes = len(np.unique(y_train))
    Y_train = np.eye(num_classes)[y_train]
    Y_test = np.eye(num_classes)[y_test]
    
    return X_train, Y_train, X_test, Y_test, y_test

# --- STEP 2: Green-ELM Hidden Layer Expansion ---
def green_elm_forward_map(X, W1, b1):
    """
    Applies the random non-linear mapping (Cover's Theorem Expansion)
    X: Shape (N, input_dim)
    W1: Shape (input_dim, hidden_dim)
    b1: Shape (1, hidden_dim)
    """
    # Linear projection followed by ReLU non-linearity
    H = np.dot(X, W1) + b1
    return np.maximum(0, H) # ReLU Activation

# --- STEP 3: Analytical Solvers ---
def solve_green_elm(H, Y, variant='cholesky', lam=1e-3):
    """
    Solves for output weights W2 analytically using specified variant.
    H: Hidden matrix (N, d)
    Y: Target one-hot labels (N, num_classes)
    """
    N, d = H.shape
    start_time = time.time()
    
    if variant == 'pinv':
        # Exact SVD-based Minimum Norm Solution
        W2 = np.linalg.pinv(H) @ Y
        
    elif variant == 'lud':
        # LU Decomposition Alternative using Normal Equations
        # Left-multiply by H.T to create a square matrix system
        A = np.dot(H.T, H) + lam * np.eye(d) # Add ridge identity for stability
        B = np.dot(H.T, Y)
        W2 = np.linalg.solve(A, B) # Uses LU factorization under the hood
        
    elif variant == 'cholesky':
        # Fast Positive-Definite Cholesky Solver
        A = np.dot(H.T, H) + lam * np.eye(d)
        B = np.dot(H.T, Y)
        # Compute Lower triangular matrix: A = L * L.T
        L = np.linalg.cholesky(A)
        # Solve L * Y_tmp = B via forward substitution, then L.T * W2 = Y_tmp via backward substitution
        W2 = np.linalg.solve(L.T, np.linalg.solve(L, B))
        
    else:
        raise ValueError(f"Unknown variant: {variant}")
        
    solve_time = time.time() - start_time
    return W2, solve_time

# --- EXECUTION PIPELINE ---
if __name__ == "__main__":
    # 1. Get backbone features (CIFAR-10 images reduced to 512 dimensions)
    X_train, Y_train, X_test, Y_test, y_test_raw = extract_backbone_embeddings(dataset_name='cifar10')
    
    input_dim = X_train.shape[1] # 512
    hidden_dim = 4000            # Hyper-dimensional target width (d)
    
    print(f"\nInitializing Random Layer: {input_dim} -> {hidden_dim} dimensions...")
    # Randomly sample projection weights using normal distribution
    W1 = np.random.normal(0, np.sqrt(2.0 / input_dim), (input_dim, hidden_dim))
    b1 = np.random.normal(0, 1.0, (1, hidden_dim))
    
    print("Mapping features to hyper-dimensional space...")
    H_train = green_elm_forward_map(X_train, W1, b1)
    H_test = green_elm_forward_map(X_test, W1, b1)
    
    # 2. Benchmarking Solvers
    variants = ['pinv', 'lud', 'cholesky']
    
    for variant in variants:
        print(f"\nRunning Solver Variant: [Green-ELM-{variant.upper()}]")
        
        # Train
        W2, solve_time = solve_green_elm(H_train, Y_train, variant=variant, lam=1e-3)
        print(f"-> Analytical Optimization Time: {solve_time:.4f} seconds")
        
        # Evaluate
        predictions = np.dot(H_test, W2)
        predicted_classes = np.argmax(predictions, axis=1)
        accuracy = np.mean(predicted_classes == y_test_raw) * 100
        print(f"-> Top-1 Test Accuracy: {accuracy:.2f}%")
\end{lstlisting}


\begin{lstlisting}[style=pythonstyle,
	caption={\textbf{CIFAR-10 Streaming Benchmarks:} Simulation of an edge deployment by initializing the model on a baseline subset (80\%) and feeding the remaining data in isolated, sequential streams of mini-batches (B = 250). It processes a 250-sample chunk with zero raw image replay storage for low-cost, low-memory embedded devices that can't afford to wait a second between updates.}
]
import torch
import torch.nn as nn
import torchvision
import torchvision.transforms as transforms
import numpy as np
import time

# Set random seed for reproducibility
np.random.seed(42)
torch.manual_seed(42)

# --- STEP 1: Feature Extraction using Frozen Backbone ---
def extract_backbone_embeddings(dataset_name='cifar10', batch_size=128):
    device = torch.device('cuda' if torch.cuda.is_available() else 'cpu')
    print(f"Extracting embeddings using device: {device}")
    
    # Load Pre-trained ResNet-18 and freeze it
    backbone = torchvision.models.resnet18(weights=torchvision.models.ResNet18_Weights.DEFAULT)
    backbone.fc = nn.Identity() 
    backbone = backbone.to(device)
    backbone.eval()
    
    # Standard normalization for ResNet
    transform = transforms.Compose([
        transforms.Resize(224), 
        transforms.ToTensor(),
        transforms.Normalize(mean=[0.485, 0.456, 0.406], std=[0.229, 0.224, 0.225])
    ])
    
    if dataset_name.lower() == 'cifar10':
        train_set = torchvision.datasets.CIFAR10(root='./data', train=True, download=True, transform=transform)
        test_set = torchvision.datasets.CIFAR10(root='./data', train=False, download=True, transform=transform)
    else:
        raise ValueError("Unsupported dataset. Please use 'cifar10'.")
        
    train_loader = torch.utils.data.DataLoader(train_set, batch_size=batch_size, shuffle=False, num_workers=2)
    test_loader = torch.utils.data.DataLoader(test_set, batch_size=batch_size, shuffle=False, num_workers=2)
    
    def get_features(loader):
        all_features = []
        all_labels = []
        with torch.no_grad():
            for images, labels in loader:
                images = images.to(device)
                features = backbone(images)
                all_features.append(features.cpu())
                all_labels.append(labels)
        return torch.cat(all_features), torch.cat(all_labels)
    
    print("Processing training embeddings...")
    X_train, y_train = get_features(train_loader)
    print("Processing testing embeddings...")
    X_test, y_test = get_features(test_loader)
    
    # One-hot encode targets using PyTorch operations
    num_classes = 10
    Y_train = torch.nn.functional.one_hot(y_train, num_classes=num_classes).double()
    Y_test = torch.nn.functional.one_hot(y_test, num_classes=num_classes).double()
    
    # Convert extracted features explicitly to double (float64) precision 
    X_train = X_train.double()
    X_test = X_test.double()
    
    return X_train, Y_train, X_test, Y_test, y_test

# --- STEP 2: Green-ELM Hidden Layer Expansion (PyTorch Backend) ---
def green_elm_forward_map_torch(X, W1, b1):
    """
    Applies the random non-linear mapping using GPU-accelerated PyTorch operations.
    """
    return torch.clamp(torch.addmm(b1, X, W1), min=0.0)

# --- STEP 3: Optimized Online RLS Equation Solver ---
def solve_green_elm_incremental_torch(H_new, Y_new, P_old, W2_old):
    """
    Performs a mathematically exact, closed-form streaming update using PyTorch.
    """
    B, d = H_new.shape
    
    # 1. Compute sub-dimensional innovation matrix in O(B^3) space
    shuttle_matrix = torch.eye(B, device=H_new.device) + torch.matmul(H_new, torch.matmul(P_old, H_new.T))
    
    # 2. Compute Kalman/Innovation Gain Matrix
    kappa = torch.matmul(P_old, torch.matmul(H_new.T, torch.inverse(shuttle_matrix))) # torch 1.5

    # 3. Update tracking inverse matrix for optimization
    P_new = P_old - torch.matmul(kappa, torch.matmul(H_new, P_old))
    
    # 4. Compute error residue and update weights without memory replay
    error_residue = Y_new - torch.matmul(H_new, W2_old)
    W2_new = W2_old + torch.matmul(kappa, error_residue)
    
    return W2_new, P_new

# --- EXECUTION PIPELINE ---
if __name__ == "__main__":
    # Determine processing acceleration backend
    device = torch.device('cuda' if torch.cuda.is_available() else 'cpu')
    print(f"Target execution hardware device: {device}")
    
    # 1. Extract Backbone Features
    X_train, Y_train, X_test, Y_test, y_test_raw = extract_backbone_embeddings()
    
    # Move full data tensors directly to target device memory 
    X_train, Y_train = X_train.to(device), Y_train.to(device)
    X_test, Y_test, y_test_raw = X_test.to(device), Y_test.to(device), y_test_raw.to(device)
    
    # 2. Split training pool into an Initial Batch (80%) and a Streaming Stream (20%)
    N_total = X_train.shape[0]
    N_initial = int(N_total * 0.8)
    
    X_init, Y_init = X_train[:N_initial], Y_train[:N_initial]
    X_stream, Y_stream = X_train[N_initial:], Y_train[N_initial:]
    
    # 3. Initialize Random Weight Layers directly on device memory
    W1 = torch.randn(512, 4000, dtype=torch.double, device=device) * np.sqrt(2.0 / 512)
    b1 = torch.randn(1, 4000, dtype=torch.double, device=device)
    
    # Generate hidden state activations
    H_init = green_elm_forward_map_torch(X_init, W1, b1)
    H_test = green_elm_forward_map_torch(X_test, W1, b1)
    
    print("\n--- Phase I: Generating Baseline Solver Ground State ---")
    lam = 1e-3
    if device.type == "cuda": torch.cuda.synchronize()
    t_start = time.time()
    
    A_init = torch.matmul(H_init.T, H_init) + lam * torch.eye(4000, device=device)
    B_init = torch.matmul(H_init.T, Y_init)
    
    #P_stream = torch.linalg.inv(A_init)
    P_stream = torch.inverse(A_init) # torch 1.5
    W2_stream = torch.matmul(P_stream, B_init)
    
    if device.type == "cuda": torch.cuda.synchronize()
    print(f"Base Training Complete in: {time.time() - t_start:.4f} seconds")
    
    # 4. Simulate Streaming Batches (Mini-batches of 250 arriving at the edge node)
    print("\n--- Phase II: Simulating Streaming Learning Updates ---")
    batch_size = 250
    num_stream_batches = X_stream.shape[0] // batch_size
    t_rls_total = 0.0
    
    # Warmup call to ensure CUDA caching allocations do not taint execution timing profiles
    _ = solve_green_elm_incremental_torch(H_init[:batch_size], Y_init[:batch_size], P_stream, W2_stream)
    
    for b_idx in range(num_stream_batches):
        start_pt = b_idx * batch_size
        end_pt = start_pt + batch_size
        
        H_chunk = green_elm_forward_map_torch(X_stream[start_pt:end_pt], W1, b1)
        Y_chunk = Y_stream[start_pt:end_pt]
        
        if device.type == "cuda": torch.cuda.synchronize()
        t_update_start = time.time()
        
        # Invoke properly aligned function signature matching step 3
        W2_stream, P_stream = solve_green_elm_incremental_torch(H_chunk, Y_chunk, P_stream, W2_stream)
        
        if device.type == "cuda": torch.cuda.synchronize()
        t_rls_total += (time.time() - t_update_start)
        
    print(f"Total time spent executing {num_stream_batches} streaming updates: {t_rls_total:.4f}s")
    print(f"Average time per unique mini-batch upload: {t_rls_total / num_stream_batches:.4f}s")
    
    # 5. Final Top-1 Precision Metrics Validation
    final_preds = torch.matmul(H_test, W2_stream)
    final_acc = torch.mean((torch.argmax(final_preds, dim=1) == y_test_raw).float()).item() * 100
    print(f"Post-Streaming Top-1 Test Accuracy: {final_acc:.2f}%")
\end{lstlisting}

\end{document}